\documentclass[11pt,a4paper]{article}
\usepackage[hyperref]{acl2018}
\usepackage{mathptmx} % Times font for text and math
\usepackage{latexsym}
\usepackage{amsmath}

\usepackage{verbatim} % for commenting out big sections
\usepackage{url}
\usepackage{graphicx}

\usepackage{pgfplots}
\usepackage{colortbl}
\usepackage{pgfplotstable}
\usepgfplotslibrary{statistics}

\newcommand{\0}{\hspace*{0.5em}}
\usepackage{todonotes}

\usepackage{xcolor}

\aclfinalcopy % Uncomment this line for the final submission
\def\aclpaperid{108} %  Enter the acl Paper ID here

\title{Consequences and Factors of Stylistic Differences\\in Human-Robot Dialogue}

\author{Stephanie M. Lukin$^{1}$, Kimberly A. Pollard$^{1}$, Claire Bonial$^{1}$, Matthew Marge$^{1}$,\\
\bfseries Cassidy Henry$^{1}$, Ron Artstein$^{2}$, David Traum$^{2}$ and Clare R. Voss$^{1}$\\
$^{1}$U.S. Army Research Laboratory, Adelphi, MD 20783\\
$^{2}$USC Institute for Creative Technologies, Playa Vista, CA 90094\\  {\tt stephanie.m.lukin.civ@mail.mil} \\}
\date{}

\begin{document}
\maketitle
\begin{abstract}

This paper identifies stylistic differences in instruction-giving observed in a corpus of human-robot dialogue. Differences in verbosity and structure (i.e., single-intent vs.~multi-intent instructions) arose naturally without restrictions or prior guidance on how users should speak with the robot. Different styles were found to produce different rates of miscommunication, and correlations were found between style differences and individual user variation, trust, and interaction experience with the robot. Understanding potential consequences and factors that influence style can inform design of dialogue systems that are robust to natural variation from human users.

\end{abstract}

\section{Introduction}
\label{intro}

When human users engage in spontaneous language use with a dialogue system, a variety of naturally occurring language is observed. A persistent challenge in the development of dialogue systems is determining how to handle this diversity. One strategy is to limit diversity and maximize the system's natural language understanding by training users {\it a priori} on what language and syntax is valid. However, these constraints could potentially yield inefficient interactions, e.g., the user may incur greater task and cognitive load trying to remember the proper phrasing needed by the system, worrying whether or not their speech will be understood if they do not get it exactly right. A broader approach to dealing with diversity is to develop more robust systems that can respond appropriately to different styles of language. A set of dialogue system policies that takes into account natural stylistic variations in users' speech would provide for a more nuanced, adaptable, and user-focused approach to interaction.  

\begin{figure}[t!]
\centering
\begin{small}
\begin{tabular}{|p{2.8in}|}
% exp1 alley p01 19-20
\hline
{\bf Dialogue 1: Lower Verbosity} \\ 

\0\0\0\0{\it U}: take pictures in all four directions  \\
{\it Robot}: executing... \\
{\it Robot}: done \\ \\

{\bf Dialogue 2: Higher Verbosity} \\ 
\0\0\0\0{\it U}: robot face north, take a picture, face south, take a picture, face east, take a picture \\
{\it Robot}: executing... \\
{\it Robot}: done \\ \\

{\bf Dialogue 3: Minimal Structure Style} \\ 
\0\0\0\0{\it U}: \underline{go through the other door} \\
{\it Robot}: executing... \\
{\it Robot}: done \\
\0\0\0\0{\it U}: \underline{take a picture}	\\ 
{\it Robot}: image sent \\ \\

% exp1 alley p01 7
{\bf Dialogue 4: Extended Structure Style} \\ 
\0\0\0\0{\it U}: \underline{face your starting position} and \underline{send a picture} \\
{\it Robot}: executing... \\
{\it Robot}: image sent \\
\hline
\end{tabular}
\end{small}	
\caption{Dialogues between Users (U) and a Robot, exemplifying stylistic differences \label{fig:examples}}
\vspace{-0.1in}
\end{figure}

% 1) taxonomy 
Rather than constrain users or develop a generalized dialogue system that attempts to cover all variations in the same way, we focus on analytic understanding of differences in observed language behavior, as well as possible causes of these differences and implications of misunderstanding. This work is a first step towards a more nuanced and flexible dialogue policy that can be sensitive to individual and situational differences, and adapt appropriately. This paper introduces a taxonomy of stylistic differences in instructions that humans issue to robots in a dialogue. The taxonomy consists of two classes: {\it verbosity} and {\it structure}. Verbosity is measured by number of words in an instruction. Dialogues 1 and 2 in Figure~\ref{fig:examples} show contrasting verbosity levels. Structure concerns the number of intents issued in an instruction: Minimal if it contains a single intent (Dialogue 3 in Figure~\ref{fig:examples} has two Minimal) or Extended if it contains more than one intent (Dialogue 4). Understanding stylistic differences can support the development of dialogue systems with strategies that tailor system responses to the user's style, rather than constrain the user's style to the expected input. The taxonomy is described in more detail in Section~\ref{sec:archetypes}.

% 2) applied to BL corpus
%To investigate these styles, we apply this taxonomy to a corpus of human-robot dialogue, 
We observe and analyze these stylistic differences in a corpus of human-robot direction-giving dialogue from \citet{marge2017exploring}. These styles are not unique to this corpus; they emerge in other human-robot and human-human dialogue, such as TeamTalk \citep{marge2011team} and SCARE \citep{stoia2008scare}. The corpus contains 60~dialogues from 20~participants (Section~\ref{sec:corpus}). The robot dialogue management in the corpus is controlled by a Wizard-of-Oz experimenter, allowing for the study of users' style with a fluent and naturalistic partner (i.e., with an approximation of an idealized automated system). 

% 3) misunderstanding
In Section~\ref{sec:consequences}, we investigate possible consequences and implications of these categorized styles in this corpus. We examine the relationship of style and miscommunication frequency, applying an existing taxonomy for miscommunication in human-agent conversational dialogue \cite{higashinaka2015towards} to this human-robot corpus. We explore the relationship between stylistic differences and other dialogue phenomena described in Section~\ref{sec:causes}, specifically whether:
\vspace{-0.3in}
\begin{itemize}
\item The rate of miscommunication is related to verbosity (H$_{1}$) and structure (H$_{2}$);
\vspace{-0.1in}
\item Latent user differences are related to verbosity (H$_3$) and structure (H$_4$);
\vspace{-0.1in}
\item Trust in the robot is related to verbosity (H$_{5}$) and structure (H$_{6}$);
\vspace{-0.1in}
\item Time/experience with the robot is related to verbosity (H$_{7}$) and structure (H$_{8}$).
\vspace{-0.1in}
\end{itemize}

% 5) discussion
Finally, we speculate about how knowledge of style, miscommunication, individual differences, trust, and experience might be leveraged to implement targeted and personalized dialogue management strategies and offer concluding remarks on future work (Sections~\ref{sec:discussion} and~\ref{sec:conclusion}).

% outline
% Section~\ref{sec:archetypes} introduces the taxonomy of stylistic differences in instruction-giving dialogues. We describe the corpus used to analyze these styles {\it in-vivo} in Section~\ref{sec:corpus}. Section~\ref{sec:consequences} examines the distribution of miscommunication across these styles, and Section~\ref{sec:causes} investigates potential causes of these stylistic differences. We conclude with a discussion in Section~\ref{sec:discussion} and future work in Section~\ref{sec:conclusion}.
%This outline is very helpful!
%DRT - folded the section references in the above discussion.

\section{Related Work}

% other corpora and why we use WoZ
A number of human-human direction-giving corpora exist, among them, ArtWalk \cite{liu2016coordinating}, CReST \cite{eberhard2010indiana}, SCARE \cite{stoia2008scare}, and SaGA \cite{lucking2010bielefeld}. The majority of existing analyses on these corpora focus on the vocabulary of referring expressions and entrainment. 
While variations in instruction-giving verbosity and structure are evident in these human-human interactions,
the goal of this work is to improve human-robot communication. Humans have different assumptions about how robots communicate and behave, and may speak differently to robots than they do to other humans.
We therefore chose a human-robot corpus for our style analysis that uses a Wizard-of-Oz for dialogue management. This allowed us specifically to isolate the style usage and miscommunication errors of the human partner (because the Wizard makes very few errors on the robot's end). 

% miscommunication
%Our analysis focused on a possible consequence of stylistic differences: miscommunication. 
Studies of human-robot automated systems tend to focus on the miscommunication errors of the dialogue system (i.e., the robot itself), rather than the miscommunication or style of the human partner. In conversational agents, the research focus is also primarily to categorize errors made by the agent, not the human, including errors in ASR, surface realization, or appropriateness of the response (e.g., \citet{higashinaka2015fatal,paek2000conversation}). 
The more generic task-oriented and agent-based response-level errors from \citet{higashinaka2015towards} map well to the user miscommunication in the corpus we examine, including excess/lack of information, non-understanding, unclear intention, and misunderstanding. Works that focus specifically on miscommunication from the user when interacting with a robot include those categorizing referential ambiguity and impossible-to-execute commands \cite{marge2015miscommunication}. These categories are common in the data we examine as well.

% causes of trust and time
In this analysis, we predict that trust will have an effect on stylistic variations. Factors of trust in co-present and remote human-robot collaboration has been studied with respect to engagement with the robot, and memory of information from the robot 
%, and other social influential factors 
\cite{powers2007comparing}. 
%Similarly, studies have shown that increased experience with a robot produces a shift in aspects of the interaction \cite{?}.

%%%%%%%%%%%%%%%%%%%%%%%%%%%%%%%%%%%%%%%%%%%%%%%%%%%%%%%%%%%%%%%%%%%%%%%%

\section{Stylistic Differences}
\label{sec:archetypes}

We describe two classes of stylistic differences for instruction-giving: differences in the verbosity of an instruction, and in the structure of the instruction. These styles emerge when decomposing a high-level plan or intent (e.g., exploring a physical space) into (potentially, but not necessarily) low-level instructions (e.g., how to explore the space, where to move, how to turn).

\subsection{Verbosity}
Verbosity is a continuous measure of the number of words per instruction. Compare the instruction in Dialogue 1 in Figure~\ref{fig:examples} ``take pictures in all four directions'' (6 words) with the instruction in Dialogue 2 ``robot face north, take a picture, face south, take a picture, face east, take a picture'' (16 words). Both issue the same plan (with the exception of a picture towards the west in Dialogue 2), yet Dialogue 1 condenses the instruction and assumes that the robot can unpack the higher-level plan. Dialogue 2 is more verbose and low-level, making reference to individual cardinal directions. Verbosity alone does not capture all style differences; additional categorization is needed.

%High-level, abstractive instructions do not necessarily correlate with lower verbosity. In the instruction ``go into that hallway slash room and take uh and go up against the wall that's on your right. take a picture'', the high-level intent is to take a picture of the wall in a room, which is described using many more words. 
%removed because Claire commented that this was not abstractive

\subsection{Structure of Instructions}

We define a {\it Minimal} instruction as one containing a single intent (e.g., ``turn'', ``move'', or ``request image''). A sequence of Minimal instructions often reveals the higher-level plan of the user. In Dialogue 3, the user issues a single instruction ``go through the other door'' and waits until the instruction has been completed. Upon receiving completion feedback from the robot (``executing'' and ``done'' responses), the next instruction, ``take a picture'', is issued. Compare this with Dialogue 4, where the intents ``face your starting position'' and ``send a picture'' are compounded together and issued at the same time. This is classified as an {\it Extended} intent structure: instructions that have more than one expressed intent. 
These structural definitions were first described in \citet{traum2018dialogue} to classify the composition of an instruction. In this work, we use these definitions to classify the style of the user.

\section{Human-Robot Dialogue Corpus}
\label{sec:corpus}

%wording changed from Claire's suggestion that we say the styles arose in the corpus, so we shouldn't word it $like we knew the styles would exist a priori and then found a corpus
We examine these styles in a corpus of human-robot dialogue collected from a collaborative human-robot task \cite{marge2017exploring}. The user and the robot were not co-present. The user instructed the robot in three remote, search-and-navigation tasks: a Training trial and two Main trials (M1 and M2). During Training, users got used to speaking to the robot. Main trials lasted for 20 minutes each, and users were given concrete goals for each exploration, including counting particular objects (e.g., shoes) and making deductions (e.g., if the space could be a headquarters environment).

Users spoke instructions into a microphone while looking at a live 2D-map built from the robot's LIDAR scanner. A low bandwidth environment was simulated by disabling video streaming; instead, photos could be captured on-demand from the robot's front-facing camera. To allow full natural language use, users were not 
%prompted with how to speak to the robot, 
 provided example commands to the robot,
though they were provided with a list of the robot's capabilities which they could reference throughout the trials. Well-formed instructions (unambiguous, with a clear action, end-point, and state) could be executed without any additional clarification (e.g., all dialogues in Figure~\ref{fig:examples}). The robot responded with status updates to the user to make it known when an instruction was heard and completed. When necessary, the robot requested instruction clarification. A human Wizard experimenter stood in for the robot's speech recognition, natural language understanding, and language production capabilities, which were guided by a response protocol.

\subsection{Corpus Statistics} 

The corpus contains 3,573 utterances from 20 users, totaling 18,336 words. 
1,981 instructions were issued. The least verbose instruction observed is 1 word (``stop''), and the most verbose is 59 words (mean 7.3, SD 5.8). Of the total instructions, 1,383 are of the Minimal style, and 598, Extended. A moderate, positive correlation exists between higher verbosity and the Extended style in this corpus ($r_{s} (1969) = .613,  p <$ .001)), supporting an intuition that more words would be found in Extended instructions. That this correlation is not stronger, however, may suggest that the verbosity metric is insufficient to capture critical elements of stylistic variation of structure. Number of words does not completely map onto the complexity or the ``packed'' nature of instructions. For example, the Minimal but highly verbose instruction from in  corpus ``continue down the hallway to the first entrance on the left first doorway on the left'' is 16 words, but the Extended instruction ``stop. take a picture'' is only 4. 
%really nice examples! KP

\section{Stylistic Differences and Miscommunication}
\label{sec:consequences}

A user's utterance is classified as a miscommunication if the following robot utterance is a request for clarification or indicates inability to comply; this occurred at least once in 216 (16\%) of the instructions in the corpus. We hypothesize that different instruction styles will differ in their overall rates of miscommunication, i.e.,
that miscommunication rates are related to verbosity (H$_1$) and structure (H$_2$).

If a scarcity of words leads to ambiguity or missing information, we would predict that verbosity and miscommunication rate would be negatively correlated. However, if more words simply yield more opportunities for erroneous or contradictory information, then we would predict a positive correlation between verbosity and miscommunication. We assessed this using binary logistic regression of verbosity on overall miscommunication presence (H$_1$). Results revealed that miscommunication significantly increases with verbosity (verbosity as a continuous independent variable, with model $\chi^2$ = 55.94, $p <$ .001, with Wald = 56.67, $p <$ .001, Nagelkerke $R^2$ = .06) 

If having more intents in a single instruction leaves more opportunities for mistakes, then we would predict that greater use of Extended structure would be positively related to miscommunication rates. To examine this relationship, we compared overall miscommunication rates and use of different instruction structures (H$_2$). The overall miscommunication rate for Minimal instructions is 8\%, while Extended is 18\%; we confirmed that Extended instructions were significantly more likely to have miscommunication (structure as a categorical independent variable, with Chi-square test, $\chi^2$(1,  {\it N} = 1969) = 40.91, $p <$ .001).

\subsection{Miscommunication Types}

While overall miscommunication rate differed significantly among styles, it would be useful to understand whether different styles are associated with different types of miscommunication, as this may inform what error-handing algorithms the system should favor. Following \citet{higashinaka2015towards}, we categorize miscommunication in the corpus according to Response-level and Environmental-level ontologies. Though this ontology was designed to categorize errors made by a virtual agent dialogue system, a number of categories are applicable to communicative errors committed by human users in situated dialogues. Figure~\ref{fig:errors} shows examples of user-miscommunication errors from the human-robot corpus and the robot response. Further explanation and expected relationships are below. 

{\it Response-level, Missing Information} instructions lack information required for execution (Dialogue 5). We expected that lower verbosity instructions would be more likely to lead to this type of miscommunication due to a lack of specificity,
as measured by fewer words, e.g., ``turn right'' is missing how many degrees to turn. On the other hand, we might expect that Extended instructions would be more likely to lead to this type because with multiple intent comes a compounding potential for lack of specificity. 

{\it Response-level, Unclear Instructions} are unclear due to phrasing or order of information presented (Dialogue 6). More verbose instructions were expected to be more prone to this type of miscommunication because more information, measured as words, has a higher potential to be misconstrued (e.g., it is unclear if ``north east"  is ``north" and ``east" or ``north-east"). However, increased information may provide additional context required for specification, the opposite relationship.
Due to compounding potential, we expected Extended style would
lead to more Unclear type errors.

\begin{figure}[t!]
\centering
\begin{small}
\begin{tabular}{|p{2.8in}|}
% exp1 alley p01 19-20
\hline
{\bf Dialogue 5: Response, Missing Information} \\ 

\0\0\0\0{\it U}: leave the room and go to your left \\
{\it Robot}: How far left should I go? \\ \\

{\bf Dialogue 6: Response, Unclear Instruction} \\ 
\0\0\0\0{\it U}: go into the center of the first room on your left and then take a picture at the west north east and south position \\
{\it Robot}: Can you repeat the positions you'd like photos from? \\ \\

{\bf Dialogue 7: Environment, Ambiguous Reference} \\ 
\0\0\0\0{\it U}: can you move forward to take a picture of the object \\
{\it Robot}: I'm not sure which object you are referring to. Can you describe it another way? \\ \\

{\bf Dialogue 8: Environment, Impossible Instruction} \\ 
\0\0\0\0{\it U}: move around the cone into the next room turn right ninety degrees \\
{\it Robot}: There is not enough space for me to move around the cone and enter the room ahead.  Would you like me to get as close as possible to the room? \\ \\

{\bf Dialogue 9: Environment, Misunderstand Capabilities} \\ 
\0\0\0\0{\it U}: okay turn to your right forty five degrees and turn on video camera \\
{\it Robot}: Our bandwidth will currently not support the video feed. Would you like me to send a photo? \\
\hline
\end{tabular}
\end{small}	
\caption{Dialogues between Users (U) and Robot with different miscommunications types \label{fig:errors}}
\vspace{-0.1in}
\end{figure}

{\it Environment-level, Ambiguous Reference} instructions include an ambiguous referent in the environment, potentially due to a lack of common ground (Dialogue 7). We expected that lower verbosity instructions, with less information (words) would have more Ambiguous miscommunication (e.g., ``go to the doorway'' versus ``go to the doorway furthest from you"). For Extended style, we hypothesize more Ambiguous type errors due to compounding potential.

{\it Environment-level, Impossible} instructions are impossible to execute in the physical space in terms of distance and dimension (Dialogue 8). 
We expected that overspecified instructions (higher verbosity or Extended)
might be more likely to be Impossible (e.g., in the more verbose, Extended instruction ``move up two feet, turn right ninety degrees, move forward seven feet’'', it is not possible for the robot to move 7 feet after completing the first two actions). 

{\it Environment-level, Misunderstood Capabilities} instructions are those in which the user misunderstands the robot's capabilities (Dialogue 9). We expect verbosity and structure to affect Misunderstood rates much as they affect Impossible miscommunication rates.

Logistic regression revealed that verbosity does not significantly correlate with any type of miscommunication that occurred ($\chi^2$ = 4.89, $p =$ .298). To examine this result in more detail, we conducted binomial logistic regression on each miscommunication type separately, asking, e.g., does verbosity predict whether the miscommunication is of the Ambiguous type or not? None of these results were significant.

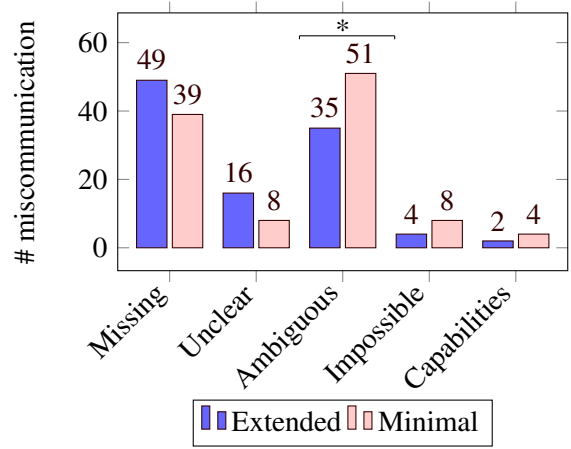
\begin{figure}
\begin{tikzpicture} 
\begin{axis}[
	xtick=data,
	ylabel=\# miscommunication, 
	enlargelimits=0.15, 
	legend style={at={(0.5,-0.50)}, 
		anchor=north,legend columns=-1}, 
	ybar, 
	symbolic x coords={Missing,Unclear,Ambiguous,Impossible,Capabilities},
	nodes near coords, 
 	nodes near coords align={vertical},
	x tick label style={rotate=45,anchor=east},
	%point meta=y % the displayed number,
	width=7.5cm,
	height=5.0cm,
	ymax=60,
	bar width=0.4cm,
] 
\addplot+[ybar, red!20!black,fill=blue!60!white,]
	coordinates {(Missing,49) (Unclear,16) (Ambiguous,35) (Impossible,4) (Capabilities,2)};
\addplot+[ybar, red!20!black,fill=red!20!white,] 
	coordinates {(Missing,39) (Unclear,8) (Ambiguous,51) (Impossible,8) (Capabilities,4)}; 

\node[above] at (200,560) {*};
%\draw (axis cs:Ambiguous,58) ++ (-15pt,0pt) -- ++(35pt,0pt);
\draw (150,600) -- (260,600);
\draw (150,600) -- (150,590);
\draw (260,600) -- (260,590);

\legend{Extended,Minimal} 
\end{axis} 
\end{tikzpicture}
\caption{\label{fig:error_type}Miscommunication types observed in structures style (* $p < 0.05$)}
\vspace{-0.1in}
\end{figure}

With regard to structure, a Chi-square test showed a non-significant trend, suggesting there may be a possible influence of structure on miscommunication type ($\chi^2$(4,  {\it N} = 216) = 8.71, $p =$ .065). We explored this result in more detail, looking at each miscommunication type separately, asking, e.g., does structure predict whether the miscommunication is of the Ambiguous type or not?  Results were significant for Ambiguous miscommunication type 
($\chi^2$(1, {\it N} = 216) = 4.01, $p =$ .045) and a trend toward significance for Unclear miscommunication type 
($\chi^2$(1, {\it N} = 216) = 3.34, $p =$ .067) With Minimal styles, miscommunications that arise are more likely to be Ambiguous type. With Extended styles, miscommunication that arise may tend to be Unclear type. Counts of miscommunication types for each structure style are shown in Figure~\ref{fig:error_type}.

\begin{figure*}
\begin{tikzpicture}
\begin{axis}
[   ybar stacked,
	enlargelimits=0.04, 
    legend style={at={(0.5,-0.2)},
    anchor=north,legend columns=-1},
    ylabel={\% of structure style}, 
    xlabel={Individual users},
    symbolic x coords={1, 2, 3, 4, 5, 6, 7, 8, 9, 10, 11, 12, 13, 14, 15, 16, 17, 18, 19, 20},
    xtick=data,
    point meta=rawy,
    height=6.8cm,
	width=15cm,
	ymax=110,
	bar width=0.46cm,
    x tick label style={rotate=45,anchor=east},
    every node near coord/.style={font=\fontsize{9}{9}\selectfont},
    nodes near coords=%
    {   \pgfkeys{/pgf/fpu,/pgf/fpu/output format=fixed}%
        \pgfmathtruncatemacro{\bigenough}{\pgfplotspointmeta > 0.05 ? 1 : 0}%
        \ifthenelse{\bigenough = 1}%
            {\pgfmathprintnumber[fixed,precision=2]{\pgfplotspointmeta}}%
            {}
        \pgfkeys{/pgf/fpu=false}%
    },
]
\addplot+[ybar, red!20!black,fill=blue!60!white,] plot coordinates {
(1,02) (2,03) (3,04) (4,11) (5,13) (6,15) (7,17) (8,26) (9,27) (10,28) (11,30) (12,40) (13,47) (14,53) (15,59) (16,63) (17,63) (18,65) (19,68) (20,71)};
\addplot+[ybar, red!20!black,fill=red!20!white,] plot coordinates {
(1,98) (2,97) (3,96) (4,89) (5,87) (6,85) (7,83) (8,74) (9,73) (10,72) (11,70) (12,60) (13,53) (14,47) (15,41) (16,37) (17,37) (18,35) (19,32) (20,29)};
\legend{{Extended},{Minimal}}
\end{axis}
\end{tikzpicture}
\vspace{-0.1in}
\caption{\label{user_structure} Percent distribution of instruction structure between users (sorted smallest to largest Extended)}
\end{figure*}
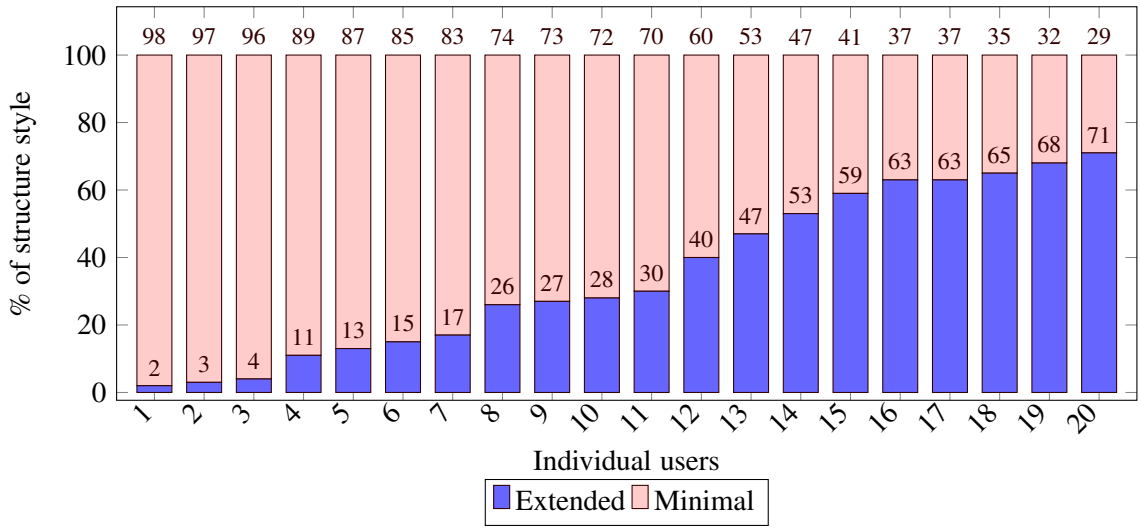

\section{Factors related to Style Differences}
\label{sec:causes}

Knowing that stylistic differences are observed in unconstrained dialogue and the relationship of these differences to miscommunication rates, it is important to assess factors of these differences in the first place. 
We examine latent individual differences, as well as trust and interaction time with the robot, which may influence style.

\subsection{Individual Differences}
\label{sec:individ}

A broad-use dialogue system can expect to receive instructions from different individuals. The dialogue system must therefore be robust to a range of individuals who will bring different speaking styles to the interaction. We hypothesized that individual users differ in their verbosity (H$_3$) and structure (H$_4$).

We first examined whether individual user identity predicted verbosity (H$_3$). The ANOVA assumption of homogeneity of variances was violated, so a Kruskal-Wallis H test was used, supporting H$_3$ with significant difference in verbosity across individual participants ($\chi^2$(19,  {\it N} = 1969) = 422.53, $p <$ .001). The most verbose user used an average of 15 words per instruction, and the least verbose used an average of 4 words.  
  
Chi-square tests revealed that individual users also vary in structure (H$_4$; $\chi^2$(19, {\it N} = 1969) = 511.70, $p <$ .001). Figure~\ref{user_structure} graphs the percentage of structural style employed by users (sorted from smallest to largest percent of Extended usage). Some users seem to simply prefer the Minimal style (Users 1, 2, 3) while other users employed a majority of Extended (Users 19, 20). Others are almost evenly split (Users 13, 14, 15).

\subsection{Trust in the Robot}
\label{sec:trust}
User trust in the robot may be a factor in how the user realizes their instructions, e.g., because the user may have different levels of confidence in the robot's abilities. Users completed the Trust Perception Scale-HRI \cite{schaefer2016measuring} after M1, and again after M2. The Trust Perception Scale-HRI is a 40-item scale designed to measure an individual's subjective perception of trust in a robot.

We hypothesized that trust in the robot would be related to verbosity (H$_5$) and structure (H$_6$). If reported trust is indicative of a user's comfort with speaking more with the robot, and/or if trust is indicative of having higher confidence in the robot's ability to process many words, then we would predict a positive relationship between trust and verbosity. On the other hand, if trust scores reflect confidence that the robot will understand instructions without need for additional words or explanations, then we would predict a negative relationship between trust and verbosity.

To assess whether and how trust levels are related to verbosity (H$_5$), we compared trust levels for a trial to the verbosity in that trial (there were not enough data points to control for individual user ID in a regression). Spearman correlation was significant, with higher trust correlating with greater verbosity 
($r_{s} (38) = .33,  p =$ .035).

If higher trust scores indicate user confidence that the robot can understand, parse, and execute complex instructions, then we predict that more Extended instructions would be observed. To assess this relationship (H$_6$), trust levels measured for each trial were compared to the proportion of Extended instructions used in that trial. 
Spearman correlation revealed a nonsignificant trend for higher trust to correlate with more use of the Extended structure ($r_{s} (38) = .29,  p =$ .07).

\subsection{Time and Experience}
\label{sec:time}

As time passes and experience grows, people are known to interact differently with technology and with communication partners.  We thus hypothesize that time/experience with the robot would be related to verbosity (H$_7$) and structure (H$_8$), i.e., as the user progresses from Training to M1 and M2, instruction-giving style may change.

If it is the case that users become more comfortable or confident as they gain more experience, we predict that verbosity should increase over time/experience (H$_7$). Indeed, verbosity increased across trials from an average of 6.1 words in Training, to 7.3 average words in M1, to 8.1 average words in MP2. A one-way repeated measures ANOVA was conducted to determine whether verbosity differed by trial (repeated measures analysis effectively controls for user ID).
Trial was significantly related to verbosity, (F(2,38) = 13.45, $ p <$ .001), and post-hoc LSD t-tests indicated that each trial had significantly more verbose instructions than previous trials (Training vs. M1 $p=$ .003; Training vs. M2 $p =$ .001; M1 vs. M2 $p =$ .020).

\begin{figure}[t!]
\begin{tikzpicture}
\begin{axis}
[   ybar stacked,
	enlargelimits=0.45, 
    legend style={at={(0.5,-0.32)},
    anchor=north,legend columns=-1},
    ylabel={\% of structure style}, 
    symbolic x coords={Training, M1, M2},
    xtick=data,
    point meta=rawy,
    height=5.8cm,
    width=7cm,
    bar width=1cm,
    x tick label style={rotate=45,anchor=east},
    every node near coord/.style={font=\fontsize{9}{9}\selectfont},
    nodes near coords=%
    {   \pgfkeys{/pgf/fpu,/pgf/fpu/output format=fixed}%
        \pgfmathtruncatemacro{\bigenough}{\pgfplotspointmeta > 0.05 ? 1 : 0}%
        \ifthenelse{\bigenough = 1}%
            {\pgfmathprintnumber[fixed,precision=2]{\pgfplotspointmeta}}%
            {}
        \pgfkeys{/pgf/fpu=false}%
    },
]
\addplot+[ybar, red!20!black,fill=blue!60!white,] plot coordinates {
({Training},22)
({M1},30)
({M2},36)
};
\addplot+[ybar, red!20!black,fill=red!20!white,] plot coordinates {
({Training},78)
({M1},70)
({M2},64)
};

% y value 
\draw (axis cs:Training,117) ++ (0pt,0pt) -- ++(40.5pt,0pt);
% center this between two bars
\node[above] at (50,880) {*};
% tiny lines going down; y values should be 3 pts apart
\draw (axis cs:Training,117) -- (axis cs:Training,114);
% this is the other tiny line. yvalues shoudl be the same as above
\draw (axis cs:M1,117) -- (axis cs:M1,114);

\draw (axis cs:Training,127) ++ (0pt,0pt) -- ++(81pt,0pt);
% center this between two bars
\node[above] at (100,980) {**};
% tiny lines going down; y values should be 3 pts apart
\draw (axis cs:Training,127) -- (axis cs:Training,124);
% this is the other tiny line. yvalues shoudl be the same as above
\draw (axis cs:M2,127) -- (axis cs:M2,124);

\legend{{Extended},{Minimal}}
\end{axis}
\end{tikzpicture}
\vspace{-0.1in}
\caption{\label{time_structure} Percent distribution of instruction structure between trials (* $p < 0.05$; ** $p < 0.01$)}
\vspace{-0.1in}
\end{figure}
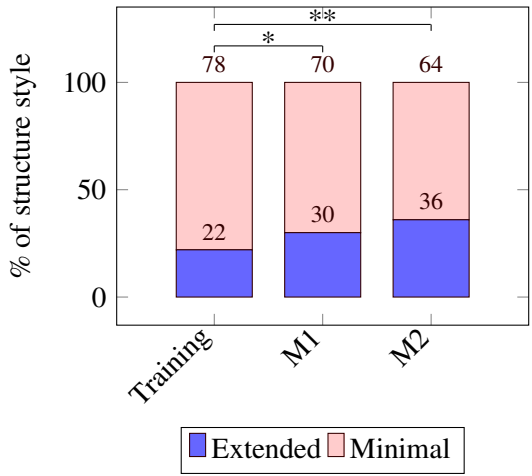

Figure~\ref{time_structure} shows the percentage of structural style in each trial. There is a general upward trend in use of the Extended style as users engage in successive trials. A one-way repeated measures ANOVA was used to determine whether structure usage differed by trial (H$_8$). Results showed significant differences among trials (F(2,38) = 8.26, $p =$ .001), with post-hoc LSD t-tests revealing greater Extended structure use in M1 and M2 as compared to Training (Training vs. M1 $p=$ .014; Training vs. M2 $p =$ .002). Structural usage between M1 and M2 was not significantly different (M1 vs. M2 $p =$ .190).

\section{Discussion}
\label{sec:discussion}

\subsection{Miscommunication}

Styles differ in the overall frequency of miscommunication they engender, but these differences are not consistent across all miscommunication types. Among miscommunication-producing instructions, we found no correlation between verbosity and what type of miscommunication was produced (H$_1$). Future analyses that look at additional linguistic features may help reveal what is happening at a level of specificity beyond a simple word count. We can speculate that this may be because user misunderstandings of the robot or environment exist regardless of how many words it takes the user to express these misunderstandings (Impossible, Misunderstood Capabilities), or because Ambiguous, Unclear, or Missing Information miscommunication can result either from too few words, or from cases where the participant adds more words and commits more miscommunication with those words. This raises the question of what it is that is being added with more verbose instructions, if not clarification information. Future research can aim to address this.

Our analyses revealed an effect of structure on miscommunication types (H$_{2}$). Minimal structure had a greater tendency to yield Ambiguous miscommunications. This may be because additional intents in an instruction offer opportunities to correct ambiguity in the first intent. For example, if the robot is told to go through the door and take a photo of the chair, the robot can use the presence or absence of a chair to settle any ambiguity about which door to go through. Without the additional intent packed into the instruction, this would remain ambiguous. 
Extended style additionally showed a nonsignificant trend toward yielding more response-level Unclear miscommunication types, which may result because Extended instructions are packed, sequentially-ordered instructions and thus have the ability to introduce miscommunication in the {\it order} of information presented. 
Missing Information, Impossible Instructions, and Misunderstood Capabilities were not significantly related to structure. These miscommunication types might not arise from the structural style, but instead stem from a fundamental misunderstanding on the user-end. 
Further analysis of the content of the instructions, rather than only the structure, may uncover if content is a factor.

\subsection{Individual Differences}

Our analysis revealed that latent differences among individuals appear to yield differences in verbosity and structure style (H$_3$ and H$_4$). Future analysis may aim to identify these latent differences. Possibilities include variations in potential for introspection, personality, perspective-taking ability, and other differences. Regardless of the underlying factors that cause individual differences, dialogue systems must be robust to a range of individuals who bring with them different stylistic tendencies.

\subsection{Varying Degrees of Trust}

We found that higher trust was related to higher verbosity. We speculate that this may be because when a user trusts in the robot's competence and capabilities, they are more likely to feel comfortable enough to speak more and be confident that the robot can parse longer instructions. Users' propensity for trust was not measured during the experimental collection, which may be an unobserved factor in this analysis. Future analysis will incorporate this additional information about users' latent traits.

\pagebreak

If trust is measured in questionnaires, or gauged by other means, this information could be incorporated as feedback for the dialogue system to appropriately adjust dialogue management strategies; as the users' trust in the robot is gauged during an interaction, the system will know to expect adjustments to verbosity and structure, so it can offer more appropriate and tailored responses to the user's style.  Furthermore, providing feedback that encourages trust (or discourages it) may be a gentle, minimally obtrusive way of guiding a user to employ a different style to avoid particular miscommunication types, if working with a less robust dialogue system.

\subsection{Effect of Interaction Time}

Users increased their verbosity (H$_7$) and use of the Extended style (H$_8$) when progressing from Training to M1 (and verbosity again when progressing from M1 to M2). We speculate that starting with lower verbosity and Minimal style during the Training trial might suggest users initially are hesitant or do not have a strong sense for the robot's language processing capabilities. Users may be learning from the training and growing in comfort level over interaction time and experience with the robot, and are willing to use more verbose or Extended instructions in successive trials. Another possible explanation might be that users face a more difficult task in the main trials as compared to training; when pressed for time in a more challenging task, users may use more words and be more prone to combine intents together. Future studies can aim to disentangle these effects.

We observed an increase in Extended style use between M1 and M2, but it did not reach statistical significance. This might suggest that any learning or strategy convergence in terms of structure that occurred from training to M1 may have mostly settled by M1.  It is possible that future work with a greater sample size will reveal that Extended style use continues to grow across trials. An understanding of interaction time or experience effects can be incorporated in the dialogue system to better support the change of user styles that emerge with repeated interactions. 

\section{Conclusion and Future Work}
\label{sec:conclusion}

This paper defines two classes of stylistic differences: verbosity and structure, and examines these styles in a corpus of human-robot dialogue with no constraints on how robot-directed instructions were formulated. We show that stylistic differences are linked to 
different rates and types of miscommunication (H$_2$), 
that latent individual differences exist (H$_3$ and H$_4$),
and that there is a relationship between style and trust (H$_5$ and H$_6$),
and style and interaction time (H$_7$ and H$_8$).

By understanding the effects of stylistic differences used in instruction-giving, we are posed to implement adjusting dialogue systems to the expectations of styles to increase user interaction and system performance. \citet{tapus2008user} has shown that users prefer a robot that tailors encouragement strategies according to users' personality (introverted or extroverted). \citet{torrey06} found that users prefer robots that tailor their speech to the human's level of expertise. We posit that dialogue systems could similarly be crafted to support and interact with different verbosity and structural styles. Future dialogue systems might adjust to the verbosity style by, for example, providing system feedback in more or less verbose styles, which may make the interaction feel more like a natural conversation. A system can adjust to the structural style by providing incremental feedback to users issuing Extended instructions to capture miscommunications early, as well as provide feedback that the system understood the compound instruction. The monitoring of trust and interaction time can be incorporated as feedback for the dialogue system to offer more appropriate responses or attentive repair strategies in advance of miscommunication being made. 

This investigation of style warrants further turn-by-turn analysis to better understand {\it where} style shift occurs during an interaction, and {\it why} particular styles are subject to increased rates of miscommunication. A future robot may be able to propose alternate courses of action for certain miscommunication types (e.g., the suggestion to offer the user a picture of the room in Dialogue 9). These propositions may be difficult for other miscommunication styles, which require contextual, environment information and specification directly from the user. Future work will investigate these alternative suggestions to study if a users' style would shift around the alternate action (e.g., reducing Minimal structure usage for Ambiguous instructions), or if the user would adapt the alternate action into their own style (e.g., continuing to use Minimal, but not repeating the same mistake).

% include your own bib file like this:
%\bibliographystyle{acl}
%\bibliography{acl2017}
\bibliography{sigdial18}

\begin{thebibliography}{}
\expandafter\ifx\csname natexlab\endcsname\relax\def\natexlab#1{#1}\fi

\bibitem[{Eberhard et~al.(2010)Eberhard, Nicholson, K{\"u}bler, Gundersen, and
  Scheutz}]{eberhard2010indiana}
Kathleen~M Eberhard, Hannele Nicholson, Sandra K{\"u}bler, Susan Gundersen, and
  Matthias Scheutz. 2010.
\newblock {The Indiana ``Cooperative Remote Search Task'' (CReST) Corpus}.
\newblock In {\em Proceedings of the International Conference on Language Resources and Evaluation\/}.

\bibitem[{Higashinaka et~al.(2015{\natexlab{a}})Higashinaka, Funakoshi, Araki,
  Tsukahara, Kobayashi, and Mizukami}]{higashinaka2015towards}
Ryuichiro Higashinaka, Kotaro Funakoshi, Masahiro Araki, Hiroshi Tsukahara,
  Yuka Kobayashi, and Masahiro Mizukami. 2015{\natexlab{a}}.
\newblock {Towards Taxonomy of Errors in Chat-oriented Dialogue Systems}.
\newblock In {\em Proceedings of the Special
  Interest Group on Discourse and Dialogue\/}. pages 87--95.

\bibitem[{Higashinaka et~al.(2015{\natexlab{b}})Higashinaka, Mizukami,
  Funakoshi, Araki, Tsukahara, and Kobayashi}]{higashinaka2015fatal}
Ryuichiro Higashinaka, Masahiro Mizukami, Kotaro Funakoshi, Masahiro Araki,
  Hiroshi Tsukahara, and Yuka Kobayashi. 2015{\natexlab{b}}.
\newblock {Fatal or Not? Finding Errors that Lead to Dialogue Breakdowns in
  Chat-Oriented Dialogue Systems}.
\newblock In {\em Proceedings of Empirical Methods for Natural Language Processing\/}. pages 2243--2248.

\bibitem[{Liu et~al.(2016)Liu, Tree, and Walker}]{liu2016coordinating}
Kris Liu, Jean E~Fox Tree, and Marilyn~A Walker. 2016.
\newblock C{oordinating Communication in the Wild: The Artwalk Dialogue Corpus
  of Pedestrian Navigation and Mobile Referential Communication}.
\newblock In {\em Proceedings of the International Conference on Language Resources and Evaluation\/}.

\bibitem[{L{\"u}cking et~al.(2010)L{\"u}cking, Bergmann, Hahn, Kopp, and
  Rieser}]{lucking2010bielefeld}
Andy L{\"u}cking, Kirsten Bergmann, Florian Hahn, Stefan Kopp, and Hannes
  Rieser. 2010.
\newblock {The Bielefeld Speech and Gesture Alignment Corpus (SaGA)}.
\newblock In {\em Proceedings of the International Conference on Language Resources and Evaluation workshop: Multimodal corpora--advances in capturing, coding and analyzing multimodality\/}.

\bibitem[{Marge et~al.(2017)Marge, Bonial, Foots, Hayes, Henry, Pollard,
  Artstein, Voss, and Traum}]{marge2017exploring}
Matthew Marge, Claire Bonial, Ashley Foots, Cory Hayes, Cassidy Henry, Kimberly
  Pollard, Ron Artstein, Clare Voss, and David Traum. 2017.
\newblock {Exploring Variation of Natural Human Commands to a Robot in a
  Collaborative Navigation Task}.
\newblock In {\em Proceedings of the First Workshop on Language Grounding for
  Robotics\/}. pages 58--66.
  
  \bibitem[{Marge and Rudnicky(2011)}]{marge2011team}
Matthew Marge and Alexander Rudnicky. 2011.
\newblock {The TeamTalk Corpus: Route Instructions in Open Spaces}.
\newblock In {\em Proceedings of the Workshop on Grounding Human-Robot Dialog for Spatial Tasks.\/}. 

\bibitem[{Marge and Rudnicky(2015)}]{marge2015miscommunication}
Matthew Marge and Alexander Rudnicky. 2015.
\newblock {Miscommunication Recovery in Physically Situated Dialogue}.
\newblock In {\em Proceedings of the Special
  Interest Group on Discourse and Dialogue\/}. pages 22--31.

\bibitem[{Paek and Horvitz(2000)}]{paek2000conversation}
Tim Paek and Eric Horvitz. 2000.
\newblock {Conversation as Action under Uncertainty}.
\newblock In {\em Proceedings of the Sixteenth conference on Uncertainty in
  artificial intelligence\/}. Morgan Kaufmann Publishers Inc., pages 455--464.

\bibitem[{Powers et~al.(2007)Powers, Kiesler, Fussell, and
  Torrey}]{powers2007comparing}
Aaron Powers, Sara Kiesler, Susan Fussell, and Cristen Torrey. 2007.
\newblock {Comparing a Computer Agent with a Humanoid Robot}.
\newblock In {\em Proceedings of Human-Robot Interaction (HRI), 2007 2nd ACM/IEEE
  International Conference on\/}. IEEE, pages 145--152.

\bibitem[{Schaefer(2016)}]{schaefer2016measuring}
Kristin~E Schaefer. 2016.
\newblock {Measuring Trust in Human Robot Interactions: Development of the
  ``Trust Perception Scale-HRI''}.
\newblock In {\em Proceedings of Robust Intelligence and Trust in Autonomous Systems\/},
  Springer, pages 191--218.

\bibitem[{Stoia et~al.(2008)Stoia, Shockley, Byron, and
  Fosler-Lussier}]{stoia2008scare}
Laura Stoia, Darla~Magdalena Shockley, Donna~K Byron, and Eric Fosler-Lussier.
  2008.
\newblock {SCARE: a Situated Corpus with Annotated Referring Expressions}.
\newblock In {\em Proceedings of the International Conference on Language Resources and Evaluation\/}. 

\bibitem[{Tapus et~al.(2008)Tapus, {\c{T}}{\u{a}}pu{\c{s}}, and
  Matari{\'c}}]{tapus2008user}
Adriana Tapus, Cristian {\c{T}}{\u{a}}pu{\c{s}}, and Maja~J Matari{\'c}. 2008.
\newblock {User-Robot Personality Matching and Assistive Robot Behavior
  Adaptation for Post-Stroke Rehabilitation Therapy}.
\newblock In {\em Intelligent Service Robotics\/} 1(2):169.

\bibitem[{Torrey et~al.(2006)Torrey, Powers, Marge, Fussell, and
  Kiesler}]{torrey06}
Cristen Torrey, Aaron Powers, Matthew Marge, Susan~R Fussell, and Sara Kiesler.
  2006.
\newblock {Effects of adaptive robot dialogue on information exchange and
  social relations}.
\newblock In {\em Proceedings of the First ACM SIGCHI/SIGART Conference on
  Human-Robot Interaction\/}. Association for Computing Machinery, pages
  126--133.
 
\bibitem[{Traum et~al.(2018)}]{traum2018dialogue}
David Traum, Cassidy Henry, Stephanie Lukin, Ron Artstein, Felix Gervits, Kimberly Pollard, Claire Bonial, Su Lei, Clare Voss, Matthew Marge, Cory Hayes, and Susan Hill. 2018.
\newblock {Dialogue Structure Annotation for Multi-Floor Interaction}.
\newblock In {\em  Proceedings of the International Conference on Language Resources and Evaluation}. 

\end{thebibliography}
\bibliographystyle{acl_natbib}

\end{document}